\documentclass[manuscript]{acmart}
\usepackage{caption}
\usepackage{subcaption}

\usepackage{xcolor}
\usepackage{hyperref}
\usepackage[normalem]{ulem} 

\AtBeginDocument{%
  }

\setcopyright{acmlicensed}
\copyrightyear{2026}
\acmYear{2026}
\acmDOI{XXXXXXX.XXXXXXX}
\acmISBN{978-1-4503-XXXX-X/2018/06}

\begin{document}

\title{When Words and Bodies Disagree: Realizing Verbal and Nonverbal Mismatch}

\author{Parisa Ghanad Torshizi}
\email{ghanadtorshizi.p@northeastern.edu}
\authornotemark[1]
\affiliation{
    \institution{Northeastern University}
    \city{Boston}
    \state{MA}
    \country{USA}
}

\author{Stacy Marsella}
\email{s.marsella@northeastern.edu}
\affiliation{
  \institution{Northeastern University}
  \city{Boston}
  \state{MA}
  \country{USA}
}

\renewcommand{\shortauthors}{}

\begin{abstract}
Traditional nonverbal behavior generation systems for virtual agents typically take utterance as input and generate behaviors that match or illustrate the spoken words. However, human nonverbal behavior is shaped by more than content of the speech. It is influenced by speaker roles, interpersonal relationships, social context, and cognitive and emotional states of interactants. Therefore, nonverbal channel carries information or emotions hidden in words. Modeling these complex verbal-nonverbal mismatches is critical for realistic virtual agents, particularly in training applications. We propose a taxonomy of mismatches, strengthening, mitigating, and leaking, and evaluate whether LLMs can select contextually appropriate mismatched behaviors. Through human-subject studies, we assess whether these context-driven behaviors, realized in a virtual human framework, produce the intended effects. Our results show that LLMs successfully select intended mismatches when prompted with social context. Furthermore, observers accurately perceive the intended directional shifts in agent’s attitude, emotions, and social function compared to matching baselines.
\end{abstract}


\begin{teaserfigure}
  \includegraphics[width=\textwidth]{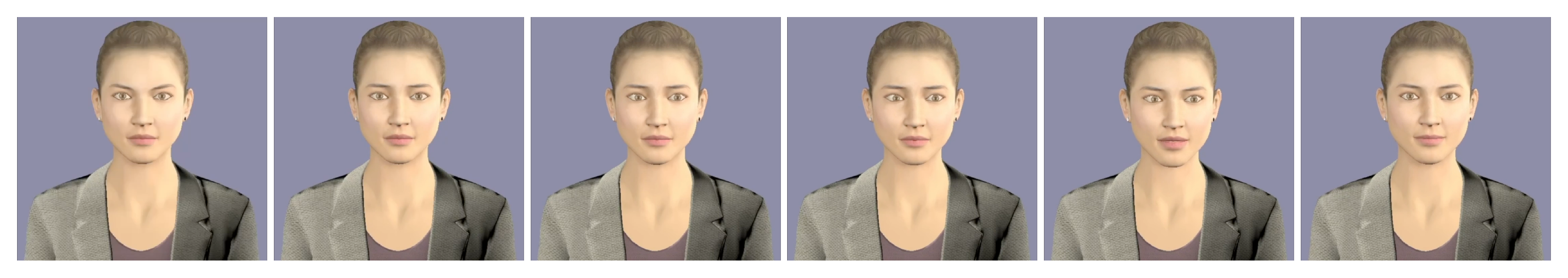}
  \caption{Frames of a virtual speaker saying "I am happy for you" with words, but communicating something else with her body.}
  \Description{This figure shows multiple frames of a virtual human diplaying the expression of sadness while uttering the sentence ''I am happy for you''. The nonverbal behaviors include a tight asymmetric smile, lowered brows, gaze directed away from the viewer, and a slight head tilt and turn.}
  \label{fig:teaser}
\end{teaserfigure}


\maketitle

\section{Introduction}
Nonverbal behaviors play an important role in human-human interaction. They communicate information, convey emotions, leak attitudes, and influence the affective state of the speakers. They also play a vital role in defining relationships and managing interactions \cite{burgoon1996nonverbal}. Their importance makes the generation of nonverbal behaviors a key agenda in the virtual human and social robotics research.
Embodied social agents are now being deployed across many different applications, including healthcare and clinical trainings, 
education, workspace and interview training, marketing, entertainment, and gaming. 

Virtual humans, in particular, are often designed to generate realistic and appropriate nonverbal behavior in line with how humans interact with each other. 
However, automated nonverbal generation approaches are often driven by the utterance being spoken\cite{nyatsanga2023comprehensive}; the model generates a set of nonverbal behaviors that tend to illustrate or repeat what the virtual human is saying (the content of the verbal channel). Although this can produce coherent behaviors, they are limited in their ability to generate richer and more nuanced relations between verbal and nonverbal behavior. 

In actual human-human interaction, nonverbal behaviors are far more expressive than simply repeating the verbal.
Ekman \cite{ekman1969repertoire} identified five semantic relationships between verbal and nonverbal channels: repeating, augmenting, illustrating, accenting, and contradicting. Although most nonverbal systems have focused on illustrating and repeating the verbal content, the contradiction relationship is comparatively underexplored. The mismatch between the verbal and the nonverbal channels does not imply a communicative failure. Rather, it is a meaningful phenomenon that carries significant social and communicative information. The communicative intent of the speaker is beyond what each channel is conveying on its own. When a person verbally communicates something positive, their nonverbal behaviors may leak hidden negative emotions, or when a person verbally communicates criticism, their nonverbal behaviors can remain friendly, warm, and open, for example to maintain the relationship.

Additionally, nonverbal behaviors are under less conscious control than dialogue and, therefore, are often more indicative of one's underlying emotions and attitudes. An example of their significance is in clinical settings \cite{philippot2003role}, where nonverbal cues may be highly informative of a patient's internal state, psychological processes, and emotional regulation.  People make judgments about situations without knowing why, or feeling emotions like distress or anxiety which they cannot control or regulate. These emotions can be expressed nonverbally, unconsciously, and be very difficult to translate into verbal expression \cite{philippot2002travail}. 

The ability to generate such complex nonverbal behaviors has direct implications on the effectiveness of virtual humans. 
Consider that virtual agents are increasingly being designed to train therapists and counselors by taking the role of a virtual patient. In such contexts, the authenticity of the virtual agent's nonverbal behavior is critical to the training outcome. A virtual patient that suppresses distress while verbally claiming to be fine or that masks emotional pain behind a composed exterior presents a far more realistic and pedagogically valuable interaction than one whose nonverbal behavior simply mirrors its words. Such nuanced behaviors allow trainee counselors to practice inferring a patient's unregulated emotional state from nonverbal cues alone, a skill that is essential in real clinical practice. Beyond clinical training, similar demands arise in negotiation training, social skills development, and deception detection scenarios, where the complexity of the interaction cannot be captured without the ability to generate mismatching verbal and nonverbal behaviors\cite{gratch2016benefits}.


Our focus in this paper is to explore automated, LLM-based approaches to selecting nonverbal behavior that realize more complex, flexible relations between verbal and nonverbal behavior, based on the interaction context, communicative intent, and mental states. The work builds on recent trends in the use of LLMs to select nonverbal behavior consistent with the role of the speaker and the context of interaction \cite{hensel2023large, torshizi2025large}, as well as personality \cite{han2025can}.
This paper continues that thread, by looking into the question of how to design the nonverbal behavior generation system to enable it to generate richer nonverbal behaviors, those with more complex and layered communicative intents. 
 
This raises another question in the design of virtual humans. What should the interface between the cognitive, emotional and dialog layers of the virtual human and its nonverbal behavior generation system be? Although current systems often only pass the utterance\cite{nyatsanga2023comprehensive}, if the situational context, relational dynamics, and social function of the interaction all shape nonverbal behavior, at times discordant with the words being spoken, then the interface must carry more than words. 

To explore this issue in the context LLM behavior selection, we developed and evaluated different prompting approaches. The goal is to assess how well they can achieve such complex nonverbal behaviors. Note that our focus is on the capability to specify such rich behavior. We are neutral to the approach of how the animation of the behavior is crafted, whether it is procedurally or data-driven \cite{nyatsanga2023comprehensive}. 

Drawing on social psychology research, we discuss a taxonomy of categories where the verbal and nonverbal behavior mismatch can occur. Using these categories, we explore different approaches to realizing these complex behaviors. Specifically, we explore alternative approaches to using large language models (LLMs), in terms of whether they can select mismatched verbal and nonverbal behaviors relevant to a provided dialog and social interaction context. We then evaluate the results in human subject studies to assess whether the resulting context-driven nonverbal behavior, when realized in a virtual human, has the intended effect on observers. 
The goal is to assess how well they can achieve richer and more complex nonverbal behavior. And, as a consequence, thereby help determine the relation, the API, between the virtual human's brain, utterance, and body.

Our contributions are as follows:
\begin{itemize}
    \item Theoretical Taxonomy: Drawing on social psychology literature, we introduce a taxonomy of verbal-nonverbal mismatches for virtual agents, organized by the social function the mismatch serves (e.g., mitigating, strengthening, leaking) and the direction of the mismatch.

    \item Technical Evaluation and Interface Design: We analyze the informational requirements for LLMs to generate appropriate mismatched behaviors. By evaluating different prompting approaches, we demonstrate that prompting the behavior generator to dialog and situational context, rather than just the utterance, eliminates the need for an external module to dictate when verbal-nonverbal mismatch should occur, suggesting a better API for virtual human architectures.

    \item Empirical Findings: We present the results of a human-subject study showing that LLM-selected mismatched behaviors successfully shift observers' attributions of the agent in the intended direction. We further find that observers can perceptually distinguish among the mismatch functions to different degrees, with some more easily separable than others.
    
    \item Dataset: We release a novel dataset of 18 dyadic scenarios (three per mismatch case type) in which nuanced verbal-nonverbal mismatches are inferable from context but never explicitly stated, designed for reuse by the research communities. Additionally, for some of these scenarios, their LLM-selected nonverbal behaviors were realized in a virtual human framework and provided as videos in the dataset.
\end{itemize}

\section{Background}
\subsection{Verbal Nonverbal Relationships}
Ekman \cite{ekman1969repertoire} outlines several relationships between verbal and nonverbal behaviors; temporally and semantically. The nonverbal act can repeat, augment, illustrate, accent, and contradict the verbal content; it can anticipate, coincide with, substitute for, or follow the words; and it can be unrelated to the verbal behavior. 
Several studies have looked at cases where these channels do not match. Jacob et al. \cite{jacob2016effects} found that slight verbal-nonverbal incongruence creates an impression of irony. Nuber et al. \cite{nuber2018attenuated} showed this irony effect is significantly weaker in high-functioning autistic individuals compared to neurotypical people.
The nonverbal channel can also convey information not apparent in speech. Mehrabian et al. \cite{mehrabian1971nonverbal} demonstrated that in deceitful communications, participants displayed more negative-affect nonverbals, suggesting unintentional leakage of internal states through the nonverbal channel. In this paper, we specifically focus on two types of verbal-nonverbal relationships and term them as \textbf{match} and \textbf{mismatch}. Match relationship, similarly to the repeat relationship defined by Ekman, describes cases where the verbal and nonverbal contain the same information; mismatch on the other hand, describes cases where the nonverbal carries information not present in the verbal channel.

\subsection{The Evolution of Behavior Generation in Virtual Agents}

Early nonverbal generation techniques mainly employed rule-based approaches \cite{cassell2001beat} \cite{lee2006nonverbal} \cite{thorisson1996communicative} \cite{cassell1994animated} \cite{pelachaud2002embodied} where domain experts manually crafted some heuristics that map from input data to nonverbal behaviors. These approaches offer interpretability and fine-grained authorial control over the behaviors. However, they rely on extensive manual efforts to develop these heuristics and does not allow for the generation of diverse gestures in a non-deterministic manner, resulting in reduction in the expressiveness of the virtual agent.

Data-driven methods were proposed to address these limitations. Early machine learning and statistical methods captured the relationship between gestures and speech by modeling the statistical properties of the underlying gesture distributions, eliminating the need for heuristics encoded by an expert. These statistical models offer more flexibility compared to the rule-based systems, by capturing more variation of nonverbal behaviors. However, they are still limited by relying on a small number of features extracted from manually annotated data, which is both labor-intensive and time consuming. Several studies have applied statistical methods to gesture generation in different ways. Some work developed gesture profiles to learn individual gesturing styles \cite{kipp2005gesture}, \cite{neff2008gesture}, while others employed Bayesian decision networks to model iconic gestures \cite{bergmann2009increasing} or nonverbal behaviors \cite{de2019non}. Elsewhere, \cite{yang2020statistics} proposed a statistical motion-graph-based system for generating body motions, and \cite{jonell2020let} used a probabilistic model to generate facial expressions. Deep learning models provide even more flexibility compared to the statistical models. Deep learning generative models aim to learn a mapping from utterances to body motions. Although these models improve the diversity and flexibility of the nonverbal behaviors, they often fall short of generating behaviors that are semantically rich. Additionally, they do offer the designer of the agent, a small amount of control over the generated behavior outputs \cite{nyatsanga2023comprehensive}.
Several work has adopted LSTMs and GANs \cite{habibie2021learning} to  generate synthesize body, hand, and face animations. \cite{yoon2020speech} proposes an architecture for gesture generation using text, audio, and speaker identity as inputs. Recent advances in nonverbal generation have increasingly centered on diffusion-based generative models and large language models (LLMs). DiffMotion \cite{zhang2023diffmotion} combines an LSTM with a diffusion model to produce gestures, DiM-Gesture \cite{zhang2024dim} employs a Mamba-2 diffusion architecture to enable personalized full-body motion generation. Several other works \cite{mughal2024convofusion, deichler2023diffusion} adopt diffusion-based approaches for multi-modal gesture synthesis. DiffuseStyleGesture \cite{yang2023diffusestylegesture} introduces a speech-driven method that integrates cross-local attention and self-attention within the diffusion pipeline. \cite{sun2025cosh, he2024co} tackle the direct generation of gesture videos from audio. At the intersection of diffusion models and LLMs, SIGGesture \cite{cheng2024siggesture} combines rhythmic gesture generation with LLM-driven semantic augmentation, while \cite{chen2025motion} leverages LLMs to jointly interpret speech and motion for controlled co-speech gesture synthesis. Similarly, \cite{pang2025llm} utilizes LLMs to achieve controllable, stylized co-speech gesture generation. 

To address the challenges of data driven approaches, \cite{hensel2023large} evaluated the use of LLMs to select gestures, and showed that LLMs are capable of selecting gestures that semantically rich and appropriate to the role of the speaker and the context of the interaction. \cite{torshizi2025large} evaluated alternative prompting approaches for gesture selection and showed that LLM-based gesture selection systems provide a high degree of control to the designer of the virtual agents. \cite{torshizicomparative} further evaluated the gesture selection abilities of multiple LLMs, to achieve a real-time gesture-selection interactive system. \cite{han2025can} explored the LLMs capability to suggest nonverbal behaviors appropriate to the speaker's personality traits.

\section{A Design Space of Verbal–Nonverbal Mismatch}
The space of verbal–nonverbal mismatch is highly complex, allowing for multiple approaches in organizing it. In this section, we propose a taxonomy organized around two primary dimensions: intentionality (whether the mismatch is intentional or unintentional) and social function (the communicative goal driving the mismatch). Based on these dimensions, we identify three distinct functions of mismatched behavior: Strengthening, Mitigating, and Leaking.\\

\textbf{Strengthening:} The speaker intends the mismatch to be interpreted by the addressee as part of the message. The goal of the mismatch is to strengthen the communicative intent of the speaker. Research on irony shows that speakers actively mark their ironical statements, intentionally informing the addressee that their statement should be interpreted as ironical; using "paracommunicative" signals, where the display of some behaviors, alongside with the ironical statement, leads the addressee to interpret the statement is ironical. For example, a statement of enthusiasm can be accommodated with a expression of boredom, or the use of intense nodding or raising of eyebrows can indicate a possibility of irony \cite{attardo2003multimodal}. The literature identifies two types of irony, Kind Irony and Sarcastic Irony \cite{anolli2000irony}.
In kind irony, the speaker uses word of scorn with intention of commendation, strengthening the positiveness of the intended message. In Sarcastic Irony, on the other hand, the speaker expresses the intention of scorn by words of commendation, strengthening the negativeness of the intended message.\\

\textbf{Mitigating:} The speaker intends to mitigate or soften the valence of their spoken words, by using mismatching nonverbal behaviors. One type of Mitigation, the Mitigated Criticism, is a form of  facework introduced by Goffman \cite{goffman1955face}: the criticism is sincerely meant, and affiliative nonverbals provide positive-politeness that compensates for the face-threatening act \cite{brown1987politeness, fletcher2016facework,trees1998managing, fraser1980mitigation}; muting the perceived aggression \cite{dews1995muting}. Conversely, we conceptualize the Mitigated Compliment, another from of Mitigation, the speaker intends to downplay the positiveness of their words \cite{bavelas1990equivocal}.

In both mitigating and strengthening functions, the mismatch is intentional; the speaker is actively trying to influence the listener's mental model of their spoken words, guiding the listener to interpret the words in the speaker's intended way. \\

\textbf{Leaking:} The speaker doesn't want to reveal their feelings but fails. Research on deception and display management demonstrates that while speakers actively work to align their displayed behavior with their spoken words, observable mismatches can arise as a failure of emotional regulation. The genuinely felt state escapes through involuntary motions, such as micro-expressions or displays in less-controllable channels \cite{ekman1985telling}. Ekman \cite{ekman1969nonverbal} establishes the leakage mechanism, and we extend it to the leakage of negative and negative emotions. Negative leakage occurs when suppressed distress, anger, or anxiety breaks through a calm verbal, whereas positive leakage occurs when unintended genuine amusement or relief breaks through a serious or neutral verbal. Unlike strengthening and mitigating, leakage is entirely unintentional, yet it provides the listener with critical observational cues to infer the speaker's true mental state.

\begin{table}[t]
\centering
\small
\setlength{\tabcolsep}{4pt}
\caption{A taxonomy of verbal--nonverbal mismatch, organized by their social function.}
\label{tab:designspace}
\begin{tabular}{llp{3.4cm}}
\toprule
Function & Intentionality & Mismatch Cases \\
\midrule
Strengthening & Intentional   & Sarcastic irony\newline Kind irony \\
\addlinespace
Mitigating    & Intentional   & Mitigated criticism\newline Mitigated compliment \\
\addlinespace
Leaking       & Unintentional & Negative leakage\newline Positive leakage \\
\bottomrule
\end{tabular}
\end{table}


\section{Exploratory Study: Analysis of LLM-Selected Behaviors}\label{sec:llm-analys}
The goal of this section is an exploration of an LLM's capability in selecting nonverbal behaviors that are not matching with the verbal content. See our repository for implementation details, including scenarios, prompting approaches and the code \footnote{Code: \url{https://github.com/parisatorshizi/When-Words-and-Bodies-Disagree}}\\
In our initial explorations, we realized when the model is prompted solely to select appropriate behaviors, meaning it is not prompted with the concept of verbal nonverbal mismatch nor any further contextual information, it tends to select only matching behavior. For example when the model was asked to select nonverbal behavior for the utterance ''This was a complete disaster'' and explain its choice; the model replied with:
\begin{verbatim}
    For "this was a complete disaster," appropriate nonverbal behaviors
    would include a furrowed brow, downturned mouth, and a slow head
    shake to convey disappointment and gravity, paired with a deflated
    posture or slumped shoulders to reinforce the sense of defeat. These
    cues align with the emotional weight of the statement, signaling
    distress and emphasizing that the outcome fell far short of expectations.
\end{verbatim}
note that in section \ref{sec:analysis}, we will see how differently the model replied when it was prompted with mismatch concept and situational information about the scenario.

This behaviors selection capability is not covering some of the verbal-nonverbal relationships defined by Ekman \cite{ekman1969repertoire}. This further inspired us to shift our efforts to evaluate multiple prompting approaches where the model is prompted with various levels of information about the interaction, and definition of verbal-nonverbal mismatch concept. Our goal therefore is to evaluate the best approach where the model can decide on the mismatch and select appropriate mismatching behaviors.



\subsection{Scenarios}
We authored 3 scenarios per mismatch case in the taxonomy. Each scenario is framed with the following characteristics: 1) it is a conversation between two people, 2) they have a specific relationship to each other, and 3) the conversation is happening in a specific context. All the characteristics are authored so that the scenario would likely elicit a verbal-nonverbal mismatch from the speaker.

Structure of each scenario: context paragraph, 4–to-6 turn dialog history that implicitly conveys the dynamics, and a final utterance of the target speaker.

We first authored 6 hand-written scenarios (one per case), and generated 12 more scenarios (the rest of the scenarios for each mismatch case) with Gemini 3.1 pro. Making the total of 18 scenarios, 3 per case, across 6 mismatch cases. We provided Gemini 3.1 Pro with our hand-written scenarios and instructed it to generate new scenarios in the same format. The scenarios are available in the Appendix \ref{appendix}.

\subsection{Prompting Conditions}\label{subsec:prompt}

Each scenario was presented under four prompting conditions, which differ only in the information accompanying the final utterance:

C1: Utterance. The final utterance alone. This condition imitates current nonverbal generation systems that take an utterance as input and produce behaviors from it.

C2: Utterance + Dialogue. The utterance is accompanied by the history of the dialog. So that the model can infer information from the dialog.

C3: Utterance + Dialogue + Context.  Context elaborates on the relationship between speakers, and provides a description of the situation. 

C4: Utterance + Dialogue+ Context + Case Label. The model is prompted with an explicit instruction to select mismatch and specifically which mismatch to select. This condition imitates an API where an external process decides the mismatch and its social function and informs the LLM-behavior-selection of it.

The system prompt was identical across conditions. It described that verbal and nonverbal channels may match or mismatch, instructed the model to determine which was appropriate from the information provided, and specified an output schema of some nonverbal channels:facial action units, gaze, head movement, and prosody, each with qualifiers: intensity, symmetry, duration, timing anchored to words in the utterance.
The model also reported the verbal–nonverbal relationship it had selected and the overall valence of its selected nonverbal behavior. The taxonomy itself was never provided except in C4.


All queries used Claude Sonnet 4.6 via the Anthropic API (SDK version 1.3.0) with the SDK's default sampling parameters and a token limit of 2000. Because generation is stochastic under these settings, each scenario–condition cell was sampled three times.

\subsection{Analysis}\label{sec:analysis}
Here is an example of Claude's output for the ''It was a complete disaster.'' utterance under the C3 condition:
\begin{verbatim}
    Speaker B's line reads as ironic teasing, not real criticism —
    opening night got a standing ovation, so calling it a disaster is
    playful exaggeration, probably riffing on Speaker A's worry about
    rushing the monologue. The nonverbals carry the joke: a suppressed
    smile (AU12) with slight cheek raise (AU6), a brief mock-serious
    brow furrow (AU4) on "complete disaster," sustained direct gaze, a
    slow mock-solemn nod on "from start to finish," and flat deadpan
    prosody with a slight drawl. Together they mark the affective
    contradiction that signals sarcasm — warmth and amusement under a
    straight face.
\end{verbatim}

\subsubsection{Mismatch rate per condition}
We counted how many times the model selected mismatch relationship per condition, evaluating what level of information is needed to get mismatched results from the model.

Mismatch selection increased with the information supplied to the model \ref{tab:mismatch_by_condition}. Given the utterance alone (C1), the model selected a mismatch in 4 of 18 scenarios. Adding the dialogue history (C2) raised this to 12, and adding the context description (C3) to 16. And telling the model explicitly to mismatch (C4), 18 times.  Cochran's Q confirmed that the conditions differed, Q(3) = 30.00, p < .001.\\

Pairwise comparisons tested the two contrasts of interest. C3 exceeded C1 (McNemar exact, p < .001, Holm-corrected): twelve scenarios shifted from matching to mismatching when dialogue and context were supplied, and none shifted in the opposite direction. C3 did not differ significantly from C4 (p = .50), the two conditions disagreeing on only two scenarios. Because C4 states the case type explicitly, this indicates that inference from dialog and situational context recovered nearly all of the mismatch selections available, eliminating the requirement of having an external process to determine that a mismatch was warranted. 

The direction of change was uniform across all comparisons: no scenario in which the model selected a mismatch under a lower-information condition reverted to matching when more information was supplied.

\begin{table}[t]
\caption{Mismatch selection by prompting condition. Behaviors is the mean number of selected behaviors per utterance.}
\label{tab:mismatch_by_condition}
\begin{tabular}{llrrr}
\toprule
Condition & Information & Mismatch & Behaviors \\
\midrule
C1 & Utterance only & 4/18 (22.2\%) & 6.6 \\
C2 & + Dialogue & 12/18 (66.7\%) & 6.9 \\
C3 & + Context & 16/18 (88.9\%) & 7.3 \\
C4 & + Case label & 18/18 (100.0\%) & 7.7 \\
\bottomrule
\end{tabular}
\end{table}

\subsubsection{Mismatch rate per case}
 
In the C1 condition, only in two scenarios did the model select a mismatch relationship: kind irony and negative leakage.

Adding the dialog history (C2) was sufficient for sarcastic irony, which reached 3/3, but produced only partial mismatching results for the mitigating cases. Once the context description was provided, five of the six case types reached 3/3.

Mitigated criticism scenarios did not do well as the others, resulting in a mismatch 1/3 times even with full context (C3), and resulting in 3/3 only when the case type was named explicitly (C4). This mismatch subcase is the hardest to model.

\begin{table}[t]
\caption{Mismatch selection by mismatch case. Each cell reports the number of the
three scenarios in that case for which the model selected a mismatch.
C1--C4 denote the utterance, dialogue, context, and case-label conditions.}
\label{tab:mismatch_by_case}
\setlength{\tabcolsep}{4pt}
\small
\begin{tabular}{@{}lcccc@{}}
\toprule
Case type & C1 & C2 & C3 & C4 \\
\midrule
Sarcastic irony & 0/3 & 3/3 & 3/3 & 3/3 \\
Kind irony & 2/3 & 2/3 & 3/3 & 3/3 \\
Mitigated compliment & 0/3 & 1/3 & 3/3 & 3/3 \\
Mitigated criticism & 0/3 & 1/3 & 1/3 & 3/3 \\
Negative leakage & 2/3 & 3/3 & 3/3 & 3/3 \\
Positive leakage & 0/3 & 2/3 & 3/3 & 3/3 \\
\bottomrule
\end{tabular}
\end{table}

\subsubsection{Mismatch rate per taxonomy function categories}
Grouping the case types by taxonomy function shows the following results. Strengthening and Leaking behaved identically at every information level, selecting mismatch 2/6 times from the utterance alone, 5/6 times with dialog, and 6/6 times once context was supplied. 
Mitigating was the harder function throughout, reaching 0/6 in C1, 2/6 in C2, and 4/6 in c3, and 6/6 in C4. The two shortfalls at C3 were both mitigated criticism scenarios. 

\begin{table}[t]
\caption{Mismatch selection by taxonomy function. Each cell reports the number
of the six scenarios in that function for which the model selected a mismatch.
C1--C4 denote the utterance, dialogue, context, and case-label conditions.}
\label{tab:mismatch_by_function}
\setlength{\tabcolsep}{4pt}
\small
\begin{tabular}{@{}lcccc@{}}
\toprule
Function & C1 & C2 & C3 & C4 \\
\midrule
Strengthening & 2/6 & 5/6 & 6/6 & 6/6 \\
Mitigating & 0/6 & 2/6 & 4/6 & 6/6 \\
Leaking & 2/6 & 5/6 & 6/6 & 6/6 \\
\bottomrule
\end{tabular}
\end{table}

\subsubsection{Behavior Compositions}

\begin{table}[t]
\caption{Mean number of nonverbal behaviors selected per utterance by information condition. Runs were averaged within each scenario--condition cell.}
\label{tab:behaviors_by_condition}
\setlength{\tabcolsep}{4pt}
\small
\begin{tabular}{@{}lcc@{}}
\toprule
Condition & Mean & SD \\
\midrule
Utterance only & 6.63 & 1.14 \\
+ Dialogue & 6.85 & 0.94 \\
+ Context & 7.26 & 0.84 \\
+ Case label & 7.70 & 1.00 \\
\bottomrule
\end{tabular}
\end{table}

\begin{table}[t]
\caption{Mean number of selected behaviors per utterance by channel and
prompting condition.}
\label{tab:channel_by_condition}
\setlength{\tabcolsep}{5pt}
\small
\begin{tabular}{@{}lccccc@{}}
\toprule
Condition & AU & Gaze & Head & Prosody & Total \\
\midrule
C1 Utterance only & 2.83 & 1.02 & 1.02 & 1.76 & 6.63 \\
C2 + Dialogue     & 3.06 & 1.09 & 1.02 & 1.69 & 6.85 \\
C3 + Context      & 3.26 & 1.17 & 1.02 & 1.81 & 7.26 \\
C4 + Case label   & 3.57 & 1.06 & 1.06 & 2.02 & 7.70 \\
\bottomrule
\end{tabular}
\end{table}

\begin{table}[t]
\caption{Mean number of selected behaviors per utterance by channel, for
generations in which the model selected a matching versus a mismatching
relationship.}
\label{tab:channel_by_relationship}
\setlength{\tabcolsep}{5pt}
\small
\begin{tabular}{@{}lccccc@{}}
\toprule
Relationship & AU & Gaze & Head & Prosody & Total \\
\midrule
Matching    & 2.72 & 1.07 & 1.03 & 1.78 & 6.60 \\
Mismatching & 3.41 & 1.09 & 1.03 & 1.84 & 7.37 \\
\bottomrule
\end{tabular}
\end{table}

Because the model specified behaviors in free text, we first created a vocabulary of the behaviors it produced. The channels differed a lot in variability. Action units used 13 distinct FACS codes across 687 objects. Gaze used 19 distinct strings across 234 objects, and head movement 28 strings across 222, in both cases the instances describing the same behavior were mapped into one behavior (for example, "direct eye contact," "direct gaze at Speaker A," and "gaze toward interlocutor", were all mapped to "gaze directed" behavior in the vocabulary). Prosody was different in kind: 393 objects spanned 151 distinct strings, most occurring once, because the model produced compositional descriptions such as "slow, flat, deadpan delivery with slight drawl."

\begin{figure}[h]
    \centering
    \includegraphics[width=0.8\textwidth]{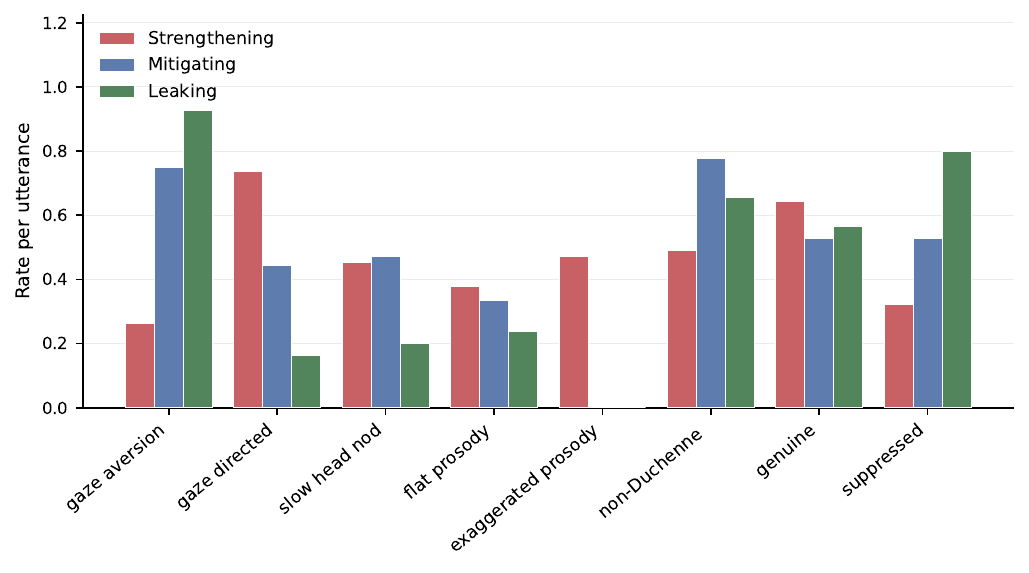}
    \caption{Rate of selected behavior features across the three taxonomy functions, for generations in which the model selected a mismatch.}
    \Description{A grouped bar chart with eight behavior features along the
    horizontal axis and rate of occurrence per utterance on the vertical axis,
    ranging from 0 to 1. Three bars per feature correspond to the strengthening, mitigating, and leaking functions. Gaze aversion is highest
    for leaking and lowest for strengthening, while directed gaze shows the
    reverse ordering with strengthening highest. Slow head nods are more
    frequent in strengthening and mitigating than in leaking. Flat prosody is
    highest in strengthening, and exaggerated prosody occurs almost exclusively
    in strengthening. The descriptor "suppressed" is most frequent in leaking
    and rare in the two intentional functions. Non-Duchenne and genuine smile
    descriptors are more evenly distributed across the three functions.}
    \label{fig:beh-comp}
\end{figure}

We therefore treated the channels differently. Gaze values were mapped to aversion or directed gaze, and head movements to nod, shake, downward tilt, lateral tilt, or turn away; the full mapping can be found in supplementary material.
We then selected some of the behaviors inside the vocabulary, based on their importance, and depicted their occurances per utterances across different mismatch function \ref{fig:beh-comp}.

As shown in Figure~\ref{fig:beh-comp}, gaze aversion is highest in the leaking condition, followed by mitigation and strengthening(Irony). Whereas direct gaze follows the opposite trend, with Irony having the highest rate. Slow head nod is higher in strengthening and mitigating ; a slow deliberate action that reveals intentional behaviors rather than unintentional (leakage). Flat and exaggerated prosody is highest in Strengthening, as argued by \cite{attardo2003multimodal} where indicated exaggerated or deadpan behaviors as signs of irony. Finally we see a distinct differences in LLM's use of the word suppressed to characterize the dynamics of the behavior, where intentional behaviors rarely used it. 


\section{Human-Subject Studies: Evaluation of LLM-Selected Behaviors}
We conducted 3 human subject studies to evaluate the LLM-generated nonverbal behaviors, using the data selected by the LLM, Claude Sonnet 4-6, for different prompting approaches. The behaviors were realized in a public domain virtual human framework \cite{Thiebaux2008} and presented as video stimuli. 
In the first two studies, we looked at the impact of the nonverbal behaviors from the perspective of conveying the valence of attitudes and emotions. Specifically, Study 1 is a comparison study evaluating whether participants perceive a difference in the virtual human's attitude between matching and mismatching verbal-nonverbal conditions. Study 2 evaluates how participants perceive the emotions the virtual human is expressing across matching and mismatching verbal-nonverbal conditions. In Study 3, we then turn to studying explicitly how effective the behaviors were in conveying the elements of our taxonomy: strengthening, mitigating, leaking.\\

The following studies are approved by University IRB board as an exempt study. All participants are recruited from Prolific. The participants in each study are mutually exclusive from the other studies. They filled out an online consent form prior to the study. And were compensated for their time through prolific, and according to Prolific's pay rates.

\subsection{Stimuli}

In the following studies, subjects are asked to assess paired match and mismatch examples of an utterance
In the human subject studies discussed in this section, we chose only one of the scenarios per each mismatch case type. We then realized its behaviors in a virtual human framework, and presented them as stimuli to the subjects.

Specifically, we realized the behaviors selected in prompting condition C1(utterance only), and prompting condition C3(dialogue + context + utterance) \ref{subsec:prompt}. The rationale behind this selection was that the C1 condition in the exploratory studies mostly selecting nonverbal behaviors that were matching the verbal content. Thus, we chose the scenarios where C1 selected strictly matching verbal-nonverbal relationships in the following studies and  we used those behaviors as the match stimuli of the studies. The mismatched stimuli were then drawn from condition 3. This is discussed further below.
Further, for Studies 1 and 2, we focused on two elements of our taxonomy: Sarcastic Irony and Mitigated Criticism. The scenarios can be found in Appendix \ref{appendix}.
It also should be noted that, for each scenario, the verbal utterance of Speaker B is constant, only the nonverbal behavior changes.\\



Specifically, in the Sarcastic Irony scenario, Speaker B's verbal utterance is positive in valence. In the match condition, the LLM-selected nonverbal behavior is also positive in valence, resulting in a positive verbal-positive nonverbal pairing, termed as \textbf{pos-pos}. In the mismatch condition, the nonverbal behavior is negative, resulting in a positive verbal-negative nonverbal pairing, termed as \textbf{pos-neg}, reflecting the Sarcastic Irony.\\

In the Mitigated Criticism scenario, Speaker B's verbal utterance is negative in valence. In the match condition, the nonverbal behavior is also negative in valence, resulting in a negative verbal-negative nonverbal pairing, termed as \textbf{neg-neg}. In the mismatch condition, the nonverbal behavior is positive and warm, resulting in a negative verbal-positive nonverbal pairing, termed as \textbf{neg-pos}, reflecting the Mitigated Criticism.

\begin{figure}[htbp]
    \centering
    \begin{subfigure}{0.4\textwidth}
        \centering
        \includegraphics[width=\textwidth]{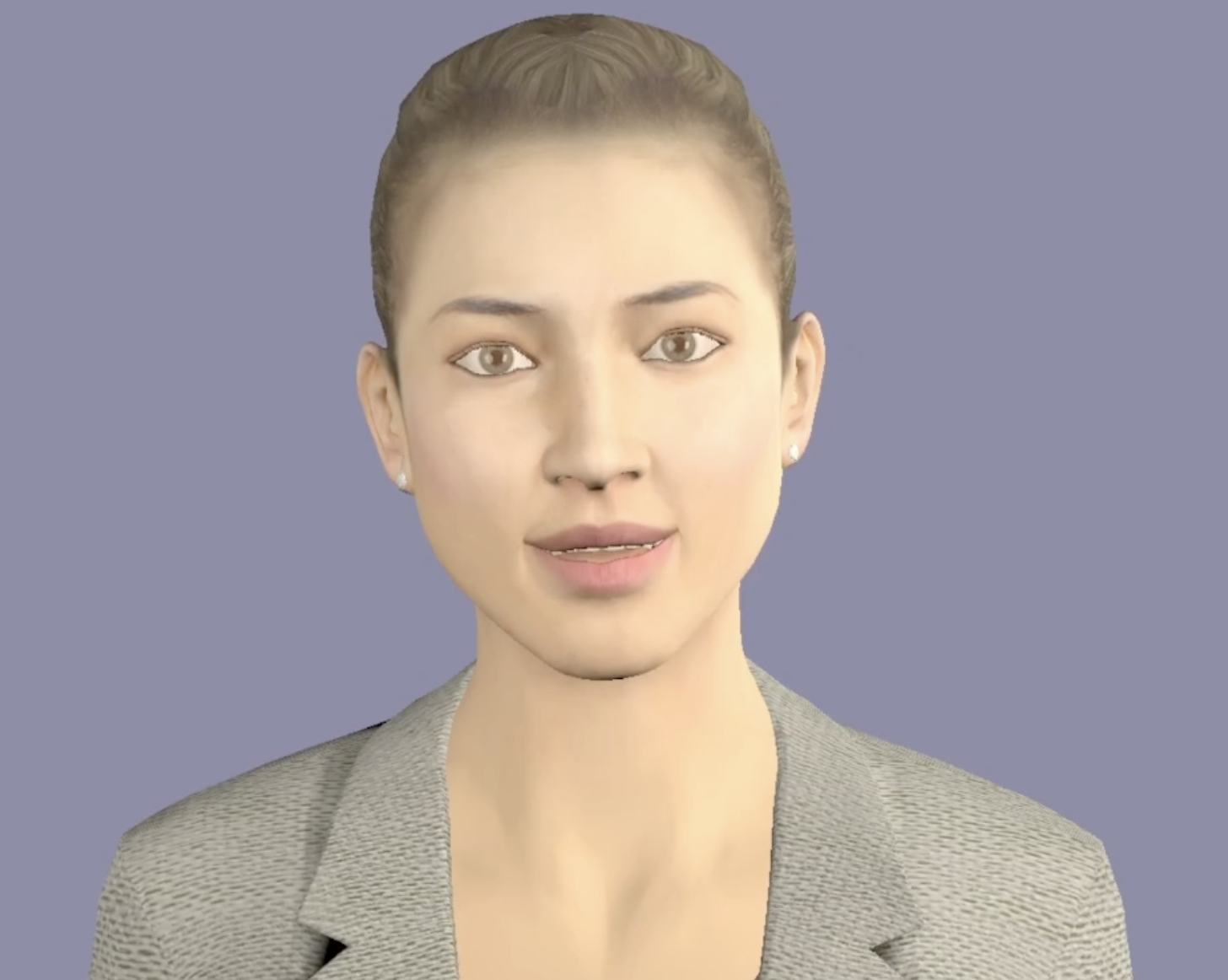}
        \caption{pos-pos:\\AU2, AU6, AU12, Head Toss}
        \label{fig:sub1}
    \end{subfigure}
    \hspace{0.05\textwidth}
    \begin{subfigure}{0.4\textwidth}
        \centering
        \includegraphics[width=\textwidth]{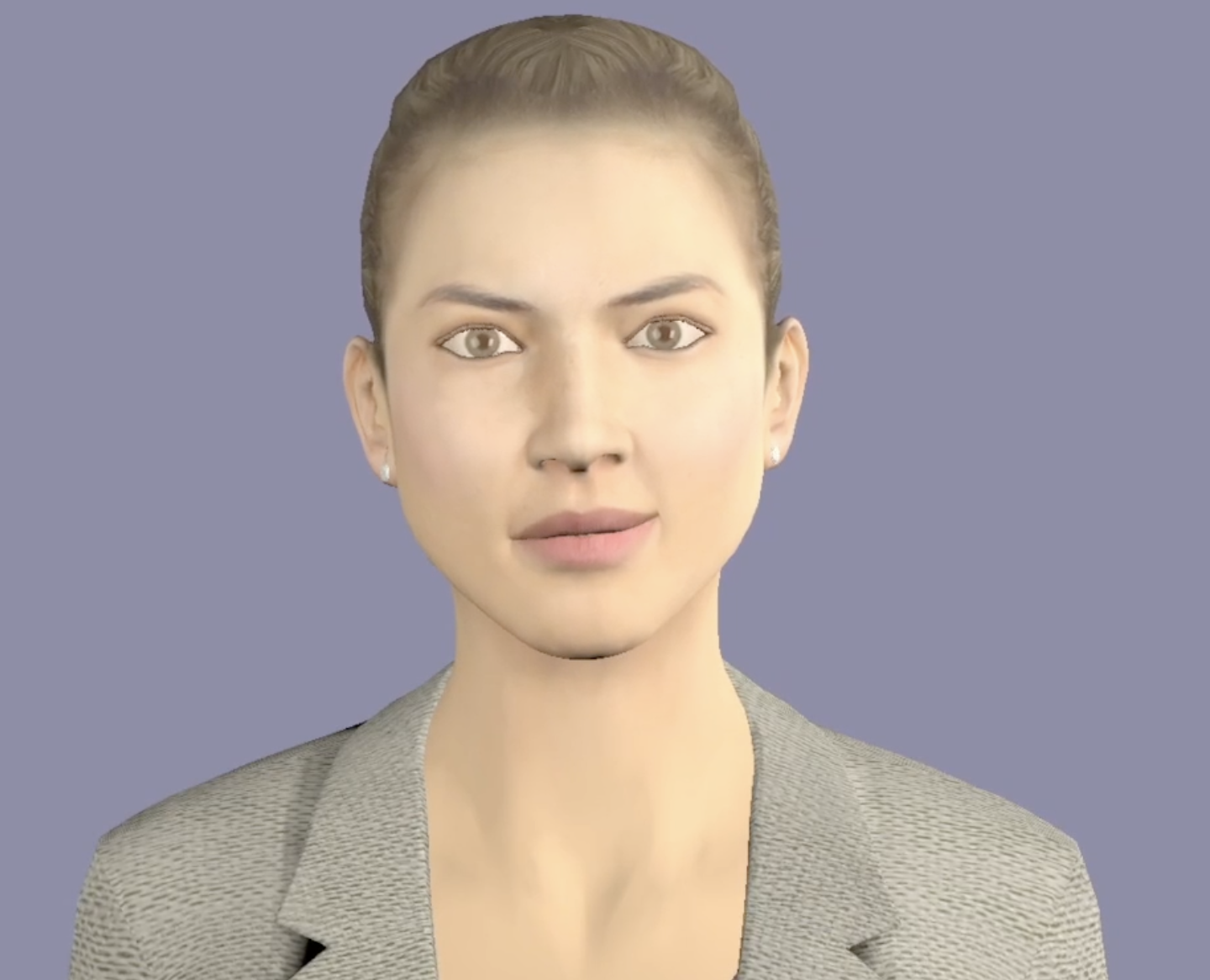}
        \caption{pos-neg:\\AU12 (Asymmetric), AU4\\}
        \label{fig:sub2}
    \end{subfigure}
    \caption{LLM-selected behavior realized in virtual human framework. The left image shows a frame of the pos-pos condition (positive verbal, positive nonverbal and match between verbal-nonverbal channels)- The right image shows the corresponding frame of the pos-neg condition(positive verbal, negative nonverbal and mismatch between verbal-nonverbal channels)}
    \Description{Two frames of the same female virtual human side by side,
    cropped to the face. In the left frame, labeled pos-pos, the brows are
    raised, the cheeks are lifted, and the mouth is drawn into a symmetric
    open smile. In the right frame, labeled pos-neg, the brows are drawn
    together and lowered and the mouth is pulled up on one side only (asymmetric smile),
    producing a one-sided smirk. The pairing shows that the asymmetry of the
    smile and the brow lowering are what distinguish the mismatching display
    from the matching one.}
    \label{fig:vh}
\end{figure}

\subsection{Virtual Human Realization}
The above mentioned LLM-selected behaviors were realized in a virtual human framework.(An example: Figure \ref{fig:vh}). The realization included post-processing; the AU amounts and the timing of the behavior relative to words and phrases in the utterance had to be mapped to corresponding parameters in the animation system. Minor additions, specifically small AU1 and AU2 activations, were applied to adjust the virtual human's default facial expression, as the LLM does not account for the character's default skeleton and facial configuration, which could otherwise cause some behaviors to appear unnatural. After realization, these stimuli were recorded as videos.


\subsection{Study 1: Perception of Attitudes}
The goal of Study 1 is to initially examine whether the LLM-selected nonverbal behavior influences the perceived attitude of a speaker when the verbal channel is held constant, across match and mismatch conditions.

\subsubsection{Study Design}
This study follows a within-subject study design. Each participant receives two Comparison sets of videos, in a randomized order. In one set, they were shown two videos (from the Sarcastic Irony scenario), the pos-pos and pos-neg, displayed side-by-side. In the other set, they were shown two videos (from the Mitigated Criticism scenario), neg-neg and neg-pos, displayed side-by-side. For each comparison set, participants were asked the following question: \emph{"Which speaker has a more positive attitude?"}, and in response, they were asked to select only one video.

\subsubsection{Hypotheses}
\begin{itemize}
    \item \textbf{H1:} LLM-selected nonverbal behaviors will shift the perception of the virtual human' attitude to be more aligned with the attitude conveyed by the nonverbal behaviors.
    \begin{itemize}
        \item H1.a: In the case where the verbal content is positive, negative nonverbals lead to more negative perception of the speaker's attitude, compared to when the nonverbals are positive.
        \item H1.b: In the case where the verbal content is negative, positive nonverbals lead to more positive perception of the speaker's attitude, compared to when the nonverbals are negative.
    \end{itemize}
\end{itemize}


\subsubsection{Participants}
49 participants were recruited via Prolific (25 female, 24 male; ages 18–54), after excluding one who did not complete the experiment. 

\subsubsection{Analysis}
The results were consistent with our hypotheses. In Comparison A, 43 out of 49 participants (87.8\%) selected the pos-pos video as having a more positive attitude, while only 6 participants (12.2\%) selected the pos-neg video. In Comparison B, 46 out of 49 participants (93.9\%) selected the neg-pos video as having a more positive attitude, while only 3 participants (6.1\%) selected the neg-neg video. Results argue that LLM-selected nonverbal behaviors effectively convey their intended attitude information, confirming H1.


In Study 2 we examine the effect of matching and mismatching verbal-nonverbal effect in more detail.

\subsection{Study 2: Perception of Emotions}
The goal of this study is to evaluate how participants perceive the feelings conveyed by the virtual human across our previously described four verbal-nonverbal conditions (pos-pos, pos-neg, neg-neg, neg-pos). Unlike Study 1, which asked participants to compare two videos, Study 2 asks participants to rate each video independently, this allowed us to evaluate how the LLM-generated nonverbal behavior impacts how the perception of virtual human's expressed emotions.

\subsubsection{Study Design}
This study follows a within-subjects design. Each participant watched all four videos (pos-pos, pos-neg, neg-neg, neg-pos) in a randomized counterbalanced order. After watching each video, participants were asked to fill out a survey on the following items:
\begin{itemize}
    \item The degree to which each of a set of emotions was being conveyed by the virtual human, using a 5-point Likert scale (1 = Strongly Disagree, 5 = Strongly Agree). This set of emotions was adopted from the Geneva Emotion Wheel (GEW) \cite{sacharin2012geneva}. From this set, 13 most relevant emotions were picked, 7 of which are positive(Amusement, Joy, Contentment, Affection, Grateful, Relief, Compassion) and the other 7 being negative(Sadness, Disappointment, Anxiety, Anger, Irritation, Contempt, Disgust). Note that we made a slight modification to the original GEW, by replacing the emotion "Love" with "Affection", and replacing the emotion "Anxiety" with "Concern".
    \item The degree to which they perceived the verbal content to be positive.
    \item The degree to which they perceived the nonverbal content to be positive.
    \item An open-ended question on their perception of what the virtual human speaker is communicating.
\end{itemize}

\subsubsection{Hypotheses}
\begin{itemize}
    \item \textbf{H2:} LLM-selected nonverbal behaviors will shift the perception of the virtual human's emotion to be more aligned with the valence of the nonverbals.
    \begin{itemize}
        \item H2.a: In the case where the verbal content is positive, negative nonverbals lead to more negative perception of the speaker's emotion, compared to when the nonverbals are positive.
        \item H2.b: In the case where the verbal content is negative, positive nonverbals lead to more positive perception of the speaker's emotion, compared to when the nonverbals are negative.
    \end{itemize}
    \item \textbf{H3:} The LLM's intended valence, for the nonverbal behavior it selected, aligns with the valence of the nonverbal behavior perceived by the participants.
\end{itemize}



\subsubsection{Participants}
39 participants were recruited via Prolific (16 female, 22 male, 1 non-binary; ages 18–54), after excluding one who did not complete the experiment. 

\subsubsection{Descriptive Statistics}
\begin{figure}[htbp]
    \centering
    \includegraphics[width=0.8\textwidth]{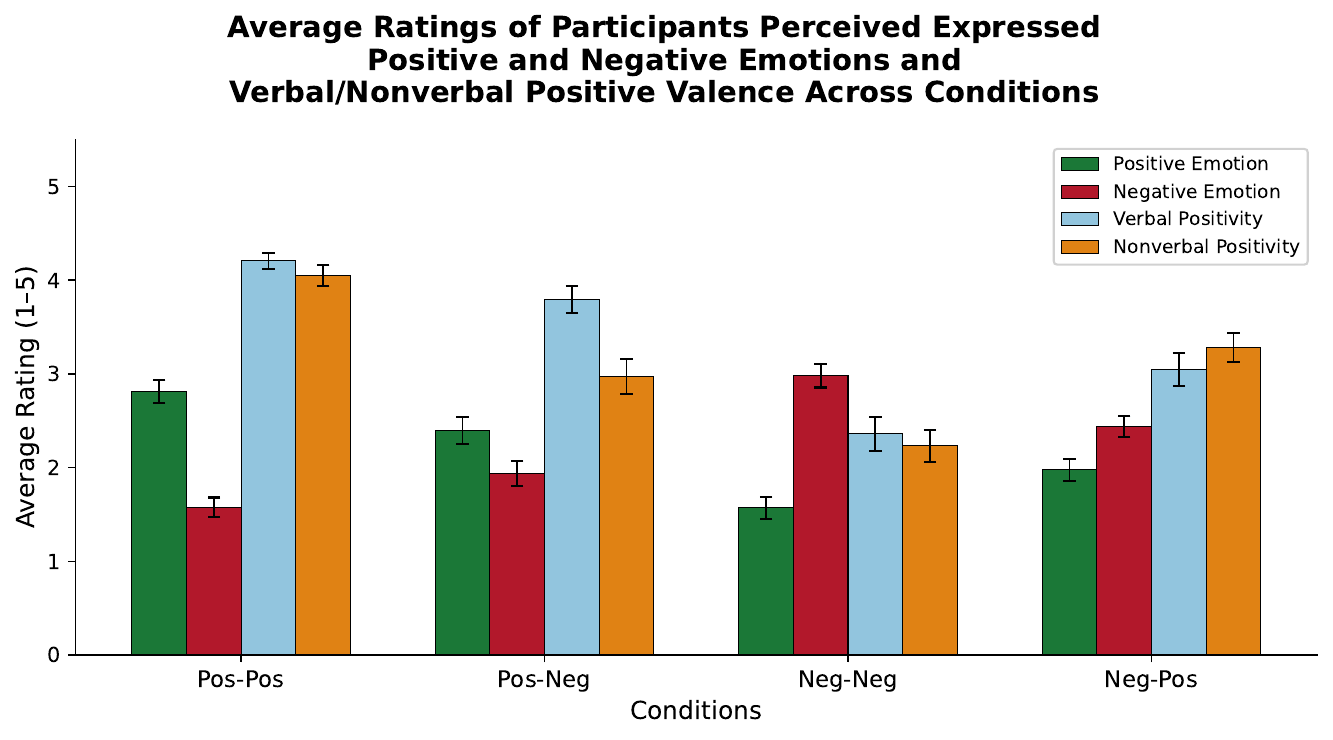}
    \caption{A demonstration of participants' perception of speaker in different conditions }
    \Description{A grouped bar chart with the four verbal--nonverbal conditions
    along the horizontal axis and mean participant rating, from 1 to 5, on the
    vertical axis. Each condition shows four bars with error bars: positive
    emotion, negative emotion, verbal positivity, and nonverbal positivity. In
    pos-pos, verbal and nonverbal positivity are both high, near 4.2 and 4.1,
    and negative emotion is lowest at 1.6. In pos-neg, nonverbal positivity
    drops to about 3.0 while verbal positivity stays high at 3.8. In neg-neg,
    both positivity measures are low, near 2.4 and 2.2, and negative emotion is
    highest at 3.0. In neg-pos, nonverbal positivity rises to about 3.3 and
    verbal positivity to about 3.1. The pattern shows nonverbal positivity
    tracking the manipulated nonverbal valence, with verbal positivity shifting
    in the same direction despite an identical verbal channel.}
    \label{fig:exp2-plot}
\end{figure}

\begin{figure}[htbp]
    \centering
    \includegraphics[width=0.8\textwidth]{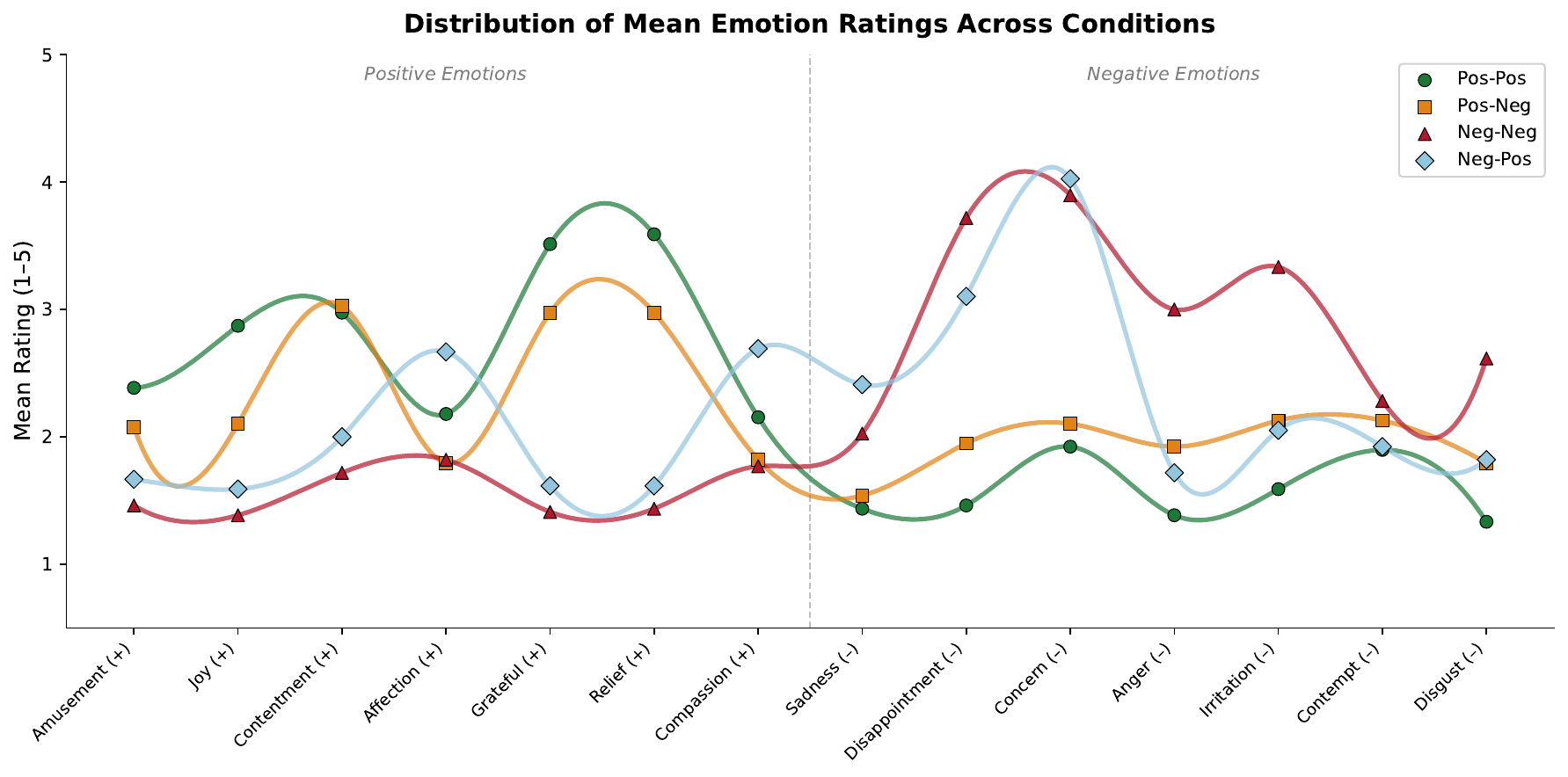}
    \caption{The distribution of speaker's expressed emotions perceived by participant's}
    \Description{A line chart with fourteen emotion labels along the horizontal
    axis and mean rating, from 1 to 5, on the vertical axis. A vertical divider
    separates the seven positive emotions on the left from the seven negative
    emotions on the right. Four lines, each with a distinct marker shape,
    correspond to the pos-pos, pos-neg, neg-neg, and neg-pos conditions. Across
    the positive emotions the two positive-verbal conditions sit above the two
    negative-verbal conditions, and this ordering reverses across the negative
    emotions. The largest separations occur at Grateful and relief among the positive emotions and at concern and disappointment among the negative emotions.}
    \label{fig:emos-plot}
\end{figure}

Positive and negative emotion composite scores were computed by averaging  the seven positive emotion items and seven negative emotion items, respectively. Table \ref{tab:descriptives}, and Figure \ref{fig:exp2-plot} report the means and standard deviation of independent variables, and Figure \ref{fig:emos-plot} report average of each emotion label across conditions.

\begin{table}[t]
\centering
\small
\caption{Descriptive Analysis of Dependent Variables Across Conditions}
\label{tab:descriptives}
\begin{tabular}{@{}lcccc@{}}
\toprule
 & \textbf{Pos Emo} & \textbf{Neg Emo} & \textbf{VB Pos} & \textbf{NVB Pos} \\
\midrule
pos-pos  & 2.81 (0.76) & 1.58 (0.66) & 4.21 (0.52) & 4.05 (0.69) \\
pos-neg  & 2.40 (0.87) & 1.94 (0.84) & 3.79 (0.89) & 2.97 (1.16) \\
neg-neg  & 1.57 (0.73) & 2.98 (0.79) & 2.36 (1.14) & 2.23 (1.09) \\
neg-pos  & 1.98 (0.73) & 2.44 (0.71) & 3.05 (1.10) & 3.28 (0.94) \\
\bottomrule
\end{tabular}
 
\vspace{4pt}
\raggedright
\footnotesize
\textit{Note.} \textit{N} = 39. Values represent means with standard 
deviations in parentheses. Scores range from 1 (\textit{Strongly Disagree}) 
to 5 (\textit{Strongly Agree}). Pos Emo = positive emotion composite; 
Neg Emo = negative emotion composite; VB Pos = Verbal behavior positive valence; 
NVB-pos = Nonverbal behavior positive valence.
\end{table}

According to Figure \ref{fig:emos-plot}, the distribution of perceived emotions vary greatly across our study conditions. To note a few, the pos-pos condition, which describes a scenario between co-workers where one delivers a report to other before the project deadline ends, is higher in "Grateful", and "Relief" emotions. The perceived emotions largely match the scenario context with a small difference between the two conditions. For those specific emotions, the pos-pos is higher in rating compared to the pos-neg, which is reasonable given that the pos-neg delivers the same verbal content of the pos-pos condition but in a sarcastic way nonverbally.\\
Another interesting point is that, the neg-pos and neg-neg conditions, which describes a scenario between a mother and a son where the mom is concerned about his son's grades, rates higher in "Disappointment" and "Concern" emotions. The perceived emotions very well match the scenario context with a small difference between the two conditions. For the "Disappointment" emotion, the neg-pos trends lower compared to the neg-neg, which is reasonable given that in the neg-pos condition the nonverbals carry positive valence and warmth to mitigate the effect of its criticism.

\subsubsection{Analysis of Variance}
A one-way repeated-measures analysis of variance (ANOVA) was conducted for each of the four dependent variables (positive emotion, negative emotion, verbal positive valence, and non-verbal positive valence) to examine the effect of condition (pos-pos, pos-neg, neg-neg, neg-pos) on participants' responses. The Greenhouse-Geisser correction was applied where the assumption of sphericity was violated. Generalized eta-squared was reported as a measure of effect size. Where the omnibus ANOVA was significant, post-hoc pairwise comparisons were conducted using estimated marginal means with a Bonferroni adjustment to correct for multiple comparisons. All analyses were conducted in R using the afex and emmeans packages.

\begin{table}[ht]
\centering
\caption{One-Way Repeated-Measures ANOVA Results for Each Dependent Variable}
\label{tab:anova}
\begin{tabular}{lcccc}
\toprule
Dependent Variable & $df$ & $F$ & $p$ & $\eta^2_G$ \\
\midrule
Positive Emotion      & 2.21, 83.82  & 40.04 & $<$.001 & .267 \\
Negative Emotion      & 2.23, 84.56  & 37.22 & $<$.001 & .336 \\
Verbal Positivity     & 2.24, 85.15  & 29.14 & $<$.001 & .365 \\
Non-Verbal Positivity & 2.66, 101.26 & 21.92 & $<$.001 & .310 \\
\bottomrule
\end{tabular}
\vspace{4pt}
\footnotesize
\\
\textit{Note.} Degrees of freedom were corrected using the Greenhouse--Geisser method. $\eta^2_G$ = generalized eta-squared.
\end{table}

A one-way repeated-measures ANOVA revealed a significant effect of condition on positive emotion, $F(2.21, 83.82) = 40.04$, $p < .001$, $\eta^2_G = .267$. Bonferroni-corrected pairwise comparisons indicated that the pos-pos condition had significantly higher positive emotion than the pos-neg condition ($p = .012$). The neg-pos condition also scored significantly higher than the neg-neg condition ($p < .001$). Consistent with the H2.a and H2.b hypotheses.
 
There was a significant effect of condition on negative emotion, $F(2.23, 84.56) = 37.22$, $p < .001$, $\eta^2_G = .336$. Pairwise comparisons revealed that the pos-pos condition had significantly lower negative emotion than the pos-neg condition ($p = .003$). The neg-neg condition had significantly higher negative emotion than the neg-pos condition ($p = .001$). Consistent with the H2.a and H2.b hypotheses.
 
A significant effect of condition on verbal positivity was found, $F(2.24, 85.15) = 29.14$, $p < .001$, $\eta^2_G = .365$. Pairwise comparisons showed that the pos-pos condition did not differ significantly from the pos-neg condition ($p = .098$). However, the neg-pos condition scored significantly higher than the neg-neg condition ($p = .002$).
 
There was a significant effect of condition on non-verbal positivity, $F(2.66, 101.26) = 21.92$, $p < .001$, $\eta^2_G = .310$. Pairwise comparisons indicated that the pos-pos condition had significantly higher non-verbal positivity than the pos-neg condition ($p < .001$). The neg-pos condition also scored significantly higher than the neg-neg condition ($p < .001$). Consistent with the H3 hypothesis.

\subsubsection{Qualitative Analysis}
Looking at human participant's free-text responses to the question as to what the virtual human is communicating, we see clear differences. below we report the keywords and some texts they used for each condition:

\begin{itemize}

\item pos-pos:
\begin{itemize}
\item Happy - no Sarcasm, Anger, Aggressiveness
\item They are trying to communicate how happy they are about the situation. 
\end{itemize}

\item pos-neg:
\begin{itemize}
\item Sarcasm, Anger, Aggressiveness
\item The speaker's words contradict her facial expression which expresses contempt and sarcasm
\item again, that some deadline was just met. though she looks angry
\end{itemize}

\item neg-pos:  
\begin{itemize}
    \item Compassion, disappointed, no anger expression
    \item The speaker's communicating both concern and compassion with her words and expression.
    \item  slight disappointment at her son's grades. 
\end{itemize}

\item neg-neg: 
\begin{itemize}
\item Anger, Disappointed, no compassion expression
\item They are angry with the person's grades. 
\item The speaker is angrily telling her son that he must work on his grades.
\item She is upset and angry about her son's grades
\end{itemize}

\end{itemize}

\subsection{Study 3: Perception of taxonomy's Social Functions}

The primary objective of Study 3 was to evaluate whether human observers could accurately perceive and interpret the dynamic nonverbal behaviors generated by the system. Specifically, we aimed to validate that the LLM-generated mismatches successfully conveyed their intended taxonomy function.

\subsubsection{Study Design}

This study follows a within-subjects design. Participants viewed the videos grouped into pairs based on the intended taxonomy function (Strengthening, Mitigating, and Leaking). Within each pair, participants watched both the Match and Mismatch videos for that specific function. The presentation order of the three functional pairs, as well as the order of the Match and Mismatch videos within each pair, was fully randomized and counterbalanced. After watching each individual video, participants were asked to fill out a survey consisting of three statements. Subjects were asked them to rate the fitness of each statement for each video in the video pairs. They rated the statements on a 5-point Likert scale (1 = Strongly Disagree to 5 = Strongly Agree). \\

It should be noted that each of the statements includes a brief description of their respective function in the taxonomy. The statements and their descriptions provided to subjects are presented below:\\

\begin{itemize}
    \item\textbf{Strengthening:} The speaker is being sarcastic or teasing.  
    \item \textbf{Mitigating:} The speaker is intentionally being indirect about what they feel, to manage listener's reaction.
    \item \textbf{Leaking:} The speaker is unintentionally revealing a feeling she did not want to show.
        
\end{itemize}

\subsubsection{Hypothesis}
To evaluate the goal of this study, we formulated two primary hypotheses:
\begin{itemize}
    \item \textbf{H4 (Directionality):} 
    Mismatching videos intended to display a function (Strengthening/mitigating/Leaking) will be rated significantly higher on their function rating (Strengthening/mitigating/Leaking) compared to their verbally-identical matching baselines. 
    \item \textbf{H5 (Discriminability):} Mismatching videos intended to display a function (Strengthening/mitigating/Leaking) will be rated significantly higher on their function rating (Strengthening/mitigating/Leaking)compared to other functions. 
\end{itemize}

\subsubsection{Participants}
57 participants were recruited via Prolific (32 female, 25 male; ages 18–54), after excluding three who did not complete the experiment.

\subsubsection{Analysis}
\paragraph{Perceptibility and Directionality of Nonverbal Mismatches}

To evaluate whether participants perceived a directional shift in the function when viewing an LLM-generated nonverbal mismatch compared to a matching baseline (H4), we conducted an Aligned Rank Transform (ART) Repeated-Measures ANOVA. The ART procedure was utilized because our ordinal 1--5 Likert rating data violated the assumption of normality, as confirmed by a Shapiro-Wilk test. Our model included \textit{Video Condition} (Matching vs. Mismatching) and \textit{Social Function} (Strengthening, Mitigating, Leaking) as within-subjects factors, with the participant modeled as a random effect. 

The ART RM ANOVA revealed a highly significant main effect of Video Condition (match and mismatch) ($F(1, 280) = 55.06, p < .001$), confirming our primary hypothesis. Overall, participants consistently rated mismatching videos higher on their intended statements than their matching counterparts. The main effect of Social Function was not significant ($F(2, 280) = 2.65, p = .072$).

The model revealed a significant Condition $\times$ Function interaction ($F(2, 280) = 4.32, p = .014$). This indicates that the magnitude of the perceptual gap between matching and mismatching behaviors varied depending on the specific taxonomy function. To analyze these differences, we utilized the ART-C procedure for multifactor contrasts \cite{elkin2021aligned}. The interaction contrasts demonstrated that the match-mismatch perceptual gap for Strengthening ($M_{match} = 1.98, M_{mismatch} = 2.99$) was significantly larger than the gap for Mitigating ($M_{match} = 2.25, M_{mismatch} = 2.63$) ($t(280) = -2.79, p = .017$). This aligns with our expectations, as strengthening behaviors (e.g., sarcastic irony) are typically more overt, whereas mitigating behaviors (e.g., polite face-saving) are inherently more subtle.

Finally, to confirm that the mismatch effect was successfully detected within \textit{every} individual function, we conducted Holm-corrected Wilcoxon Signed-Rank tests as a robustness check for the simple main effects. The results confirmed that mismatching videos were rated significantly higher than matching videos across all three taxonomic categories: Strengthening ($V = 736.5, p < .001$), Mitigating ($V = 502.5, p = .007$), and Leaking ($V = 816.5, p < .001$).

Taken together, these results demonstrate that our LLM-powered system generates perceptible, directional shifts in social meaning that human observers can reliably detect across a selection of complex social functions (H4).

\paragraph{Discriminability of Social Functions}

To explore whether observers could reliably distinguish between the three nonverbal mismatch functions (H5), we isolated the dataset to only include ratings from the mismatching condition. We employed an Aligned Rank Transform (ART) Repeated-Measures ANOVA, assigning \textit{Function} (Strengthening, Mitigating, Leaking) and \textit{Statement} (Strengthening, Mitigating, Leaking) as within-subjects factors. 

The omnibus ART RM ANOVA revealed a statistically significant interaction between the generated nonverbal function and the rated statement, $F(4, 448) = 2.48$, $p = .043$. This significant interaction confirms that observer ratings were not uniformly elevated by general behavior; rather, the ratings depended specifically on the alignment between the simulated social function and the targeted perceptual statement. 

To determine the exact boundaries of this separation, we conducted post-hoc interaction contrasts using Holm-corrected $p$-values. The results indicate varying degrees of discriminability across the taxonomy:

\begin{itemize}
    \item \textbf{Leaking (Distinct Separation):} The leaking behaviors were highly discriminable than other behaviors. When evaluating the leaking statement, the Leaking function scored significantly higher than both the Mitigating function ($p = .016$) and the Strengthening function ($p = .037$). 
    \item \textbf{Strengthening (Strong but Overlapping):} The strengthening behaviors reliably separated from Mitigating behaviors. On the strengthening statement, Strengthening video(function) scored significantly higher than the Mitigating video(function) ($p < .001$). However, Strengthening did not statistically separate from the Leaking function on this statement ($p = 1.000$), suggesting that sarcasm  occasionally was being confused with emotional leakage.
    \item \textbf{Mitigating:} The mitigating scenario was the most difficult for observers to isolate. Contrasts on the Mitigating statement did not yield significant separation from the other functions; arguing for the subtlety of the the nonverbal behaviors in the mitigation scenarios.
\end{itemize}

Overall, our LLM-behavior selection approach could successfuly generate distinct categories, though the boundaries between mitigation, sarcasm, and leakage arguably may reflect the natural, overlapping complexities of human nonverbal communication and its perception.

\section{Discussion}
This paper discusses a shift in the design of nonverbal behavior selection systems: Moving away from a common theme of systems that select nonverbal behaviors from the utterance, which  mainly results in selection of nonverbal behaviors matching to the verbal, to an approach whereby systems that infer the social context of the interaction. This approach enables the selection of more complex nonverbal behaviors, specifically the cases where the nonverbal channel mismatches the words, carrying information not apparent in the verbal channel.

To evaluate this approach, we first present a taxonomy of verbal-nonverbal mismatches into different function: Strengthening/Irony, Mitigating/face-maintenance, leaking/emotional-leakage. Then scenarios were authored  which were likely to elicit specific mismatch cases in the taxonomy.

In our first analysis \ref{sec:llm-analys}, we prompted an LLM to explore its mismatching behavior selection capabilities. We evaluated four different prompting approaches, starting from baseline (C1: utterance-only) to conditions where the mismatch is implicit in the prompt (C2: dialogue history+utterance, C3:context+ dialogue history+utterance), to explicitly instructing the LLM to select mismatching behaviors(C4: giving the model the taxonomy mismatch case directly). The results of this exploratory analysis show a trend, the more contextual information we provided in the prompts, the more times the LLM was selecting mismatching behaviors. These results argue for the importance of providing contextual information to the behavior selection system (including relationship of the speakers, and the context of their interaction). Additionally, there was a slight difference between the results of C3 and C4, meaning that providing the contextual information is producing almost similar number of mismatches as the explicit instructions to mismatch. This suggests a design of the API between the ''brain'' of the virtual human and its behavior selection. It argues that instead of a secondary system deciding the mismatch type(e.g., sarcasm) and passing it to the behavior selection system (what C4 supports), the behaviors selection system can dynamically select the behaviors based on the information provided during the interaction(what C3 supports).
In addition to number of times the LLM was selecting mismatches, the specific nonverbal behaviors it was selecting were also interesting. For example, the LLM was often selecting gaze aversion for scenarios that were intended to elicit emotional leakage or mitigation. On the other hand, directed gaze was mostly selected for Ironical scenarios.

We then proceeded to evaluate how people perceived the nonverbal behaviors selected by the LLM, which were realized in a virtual framework. this analysis was done through a series of human-subject studies.
In human subject study 1, participants compared the attitude of the virtual speaker, in matching vs mismatching conditions. The results of this study demonstrate that nonverbal behaviors were reliably interpreted as impacting the participants' interpretation of the speaker's attitude, even though the verbal content was identical.

Human subject study 2 aimed for a deeper analysis of the perception of the mismatching behaviors. Here we see that the nonverbals are reliably impacting the subject's perception of the virtual human's emotion. Behaviors selected to convey negative emotions led to perceptions of increased negativity and behaviors selected to convey positive emotions led to perceptions of increased positivity in virtual human. In addition, although not initially hypothesized, the emotional valence of the nonverbal behaviors appeared to shift how participants perceived the valence of the verbal channel itself. For example, when exposed to two videos featuring the exact same positive utterance, participants rated the spoken words as significantly more positive when paired with positive nonverbals compared to negative nonverbals. A similar trend held true for negative verbal content. 

Study 3 aimed to evaluate whether participants could accurately distinguish the different taxonomy functions from one another, and whether the mismatching videos were perceived as their intended social function compared to matching baselines. The results indicate that participants consistently perceived the intended direction of the mismatch compared to the baseline. However, the ability to distinguish the specific functions from one another was more nuanced. While behaviors like emotional leakage were highly distinguished from other behaviors, strengthening and mitigating behaviors exhibited more overlap. One potential explanation is that when the recommended behaviors, especially when they are realized in a virtual human, are hard to detect from each other. Alternatively, this also raises questions about how they were realized.

Our future work will include a more thorough analysis of this behavior selection approach across scenarios. However, we think it is critical to evaluate also in a real-time interactive applications designed for a specific domain.

\section{Limitations}

Although the LLM selected provided detailed nonverbal behaviors, mapping those behaviors into our animation system required manual mapping. We adjusted action unit amounts and timings to match the virtual human's specific skeleton. Interactive virtual human applications require real-time mapping of the behavior-selection's output into the framework, to ensure a smooth interaction. Furthermore, framework limitations prevented the realization of vocal prosody, despite the LLM suggesting it at times. And finally, such nuanced complex behaviors would benefit from more realism of the face and the voice of the virtual human, suggesting the future work of integrating this approach into diffusion model-based virtual avatars.
Another limitation is that we did not include gestures and postural shifts. This was largely due to unrelated implementation issues in the character animation system. It remains for future work to explore these other dimensions of nonverbal behavior. 

Additionally, regarding our human-subject evaluations, in study 3, participants viewed the videos without getting any information on situational context. While this was a deliberate choice to ensure participants were evaluating the nonverbal behavior rather than inferring or reasoning from a scenario context, it forces them to infer meaning from the behavior alone, which is a significantly harder test. Moreover, our evaluation is constrained by limited stimulus. Because we evaluated one video per mismatch case type in the human subject studies and relied on a single LLM, our effects are tied to these specific scenarios and model choices. We cannot claim that this exact pattern of behavior selection would hold across different model families.

Regarding the LLM, there is always a speed issue and the need for local implementation to ensure reasonable real time interactive behavior. Frankly, this is less of a concern since recent work suggests a RAG and SLM approach can often effectively address this issue.


\section{Conclusion}
In this paper, we presented an approach to nonverbal behavior selection that moves beyond rigid, utterance-driven matching to dynamic, context-aware LLM behavior selection. Through a series of human-subject studies, we demonstrated that this system successfully generates complex verbal-nonverbal mismatches that directionally influence observer perception. Our findings confirm that LLM-selected behaviors reliably shift the perceived attitude and emotions of the virtual human, effectively generating nuanced social functions like emotional leakage and mitigation. Our future work will transition this architecture from isolated perceptual evaluations into a real-time, fully interactive applications.





\section{Appendices}\label{appendix}


\begin{verbatim}
Here you can find a subset of the scenarios list that were realized in the virtual human framework:
Sarcastic Irony:
Context: Speaker A and Speaker B are colleagues of equal status working on a joint project. They have 
worked together for years and have a comfortable rapport.
Speaker B has been waiting all day for Speaker A to submit her section of the report. Speaker A finally
sends
it at 4:50, just before the 5pm deadline.
Speaker B: Hey, just checking in — did you get a chance to start on your section yet? I need it to 
put the report 
together. 
Speaker A: Yeah yeah, I\'m on it. Don\'t worry. 
Speaker B: Okay, the deadline is at 5. It\'s already 4:30. 
Speaker A: Relax, you\'ll have it.
(20 minutes passes and Speaker B shows up at Speaker A\'s office)
Speaker A: I just sent the report you were waiting for.
Speaker B: Got it, thanks. Just in time.
Kind Irony: 
Speaker B is a theater director and Speaker A is the lead actor.
They are backstage immediately following the opening night performance, 
which just ended with a standing ovation from the audience.
Speaker A: I can't believe we finally opened. The crowd was amazing.
Speaker B: They certainly made a lot of noise.
Speaker A: I know I completely rushed the pacing during that final monologue, though.
Speaker B: It was a complete disaster from start to finish.
Mitigated Criticism:
Speaker B is the mother of Speaker A, a 15-year-old boy. Speaker B has just reviewed her son's report 
cardand his grades have dropped significantly.
Speaker B: Hey, come sit with me for a second.
Speaker A: Sure, what's up?
Speaker B: I just finished going through your report card.
Speaker A: Oh. How bad is it?
Speaker B: We need to work on these grades.
Mitigated Compliment:
Speaker B is a guest at a dinner party hosted by Speaker A.
They are sitting in the living room after dessert. Speaker A has just finished performing a 
fifteen-minute, amateur acoustic guitar song for the guests.
Speaker A: That one was a little rough. I'm still trying to figure out chord progression for the bridge.
Speaker B: It sounded like a very complex piece of music.
Speaker A: Thanks. I have another one I wrote last week about my cat, if you guys don't mind me playing 
one more.
Speaker B: Hearing another one would be fine.
Negative Leakage: 
Speaker B and Speaker A are best friends who have lived in the same city for years.
They are sitting in a coffee shop. Speaker A has just unexpectedly announced they accepted a dream job 
across the  country and will be moving in two weeks.
Speaker A: They want me to start on the first of the month, so I have to fly out next weekend to find 
an apartment.
Speaker B: Wow. That is really fast. You're going to be all the way in Seattle.
Speaker A: I know it's incredibly sudden. I just couldn't pass up the opportunity.
Speaker B: I'm really happy for you.
Positive Leakage:
Speaker B and Speaker A are colleagues who have had an unspoken mutual attraction for months.
They just ran through a sudden downpour and are now squeezed tightly together under a very
small awning.
Speaker A: My clothes are completely soaked through.
Speaker B: You are the one who insisted we walk instead of calling a car.
Speaker A: Well, there is barely enough room for both of us to stand under here.
Speaker B: This is a terrible situation.


\end{verbatim}

\begin{acks}
\end{acks}

\section*{Ethics and Privacy Statement}


This work was approved by the authors' Institutional Review Board as an exempt study. Participants were recruited through Prolific, gave informed consent before participating, were compensated at Prolific's recommended rates, and could withdraw at any time. Data were anonymized; no personally identifying information beyond what Prolific requires for compensation was collected or retained. All stimuli were synthetic, dialogue written by the authors or generated with a language model and then author-curated, realized in an animated character, so no human likeness, voice, or personal data was used.
The main risk we see follows from how this methodology can be misused. We show that an large language model can be prompted to select nonverbal behavior that deliberately mismatches what an agent says, and that observers reliably read those behaviors as intended. One use case of such approach is its integration as nonverbal-behavior selection system for a virtual patient, or to make and agent better at appearing warm while communicating something serious, or at seeming to conceal a feeling it does not have. Systems built on this technique could be used to manufacture apparent sincerity, or apparent emotional leakage in commercial, persuasive, or political contexts, where a user has no way to know that an agent's expressive incongruence was authored.
We do not see privacy, fairness, or data sovereignty risks specific to this work, and note two limitations relevant to fairness. Our stimuli use a single animated character with one apparent gender and one apparent ethnicity, and our participants were English-speaking Prolific users. Nonverbal behaviors are varied across different cultures and languages, so we make no claim that either the taxonomy or the perceptual findings generalize across cultures. The scenario set is released for reuse and carries the same limitation.

\bibliographystyle{ACM-Reference-Format}
\bibliography{bib}

\appendix
.

\end{document}